\pdfoutput=1

\documentclass[11pt]{article}

\usepackage[]{emnlp2021}

\usepackage{times}
\usepackage{latexsym}
\usepackage{booktabs}
\usepackage{graphicx}
\usepackage[disable]{todonotes}
\usepackage{amsmath}
\usepackage{multirow,tabularx}

\usepackage[T1]{fontenc}

\usepackage[utf8]{inputenc}

\usepackage{microtype}

\newcommand{\xlarge}{\emph{large}}
\newcommand{\xbase}{\emph{base}}

\title{Classification-based Quality Estimation: Small and Efficient Models for Real-world Applications}

  \author{\\
  Shuo Sun,\textsuperscript{1} Ahmed El-Kishky,\textsuperscript{2} Vishrav Chaudhary,\textsuperscript{3}\\
  James Cross, \textsuperscript{3} Francisco Guzm\'an,\textsuperscript{3} Lucia Specia\textsuperscript{4}\\
  \textsuperscript{1}Johns Hopkins University, 
  \textsuperscript{2}Twitter Cortex,
  \textsuperscript{3}Facebook AI,
  \textsuperscript{4}Imperial College London\\
  \texttt{\textsuperscript{1}ssun32@jhu.edu}, \texttt{\textsuperscript{2}aelkishky@twitter.com}  \\
  \texttt{\textsuperscript{3}\{vishrav,jcross,fguzman\}@fb.com}, 
  \texttt{\textsuperscript{4}l.specia@imperial.ac.uk}\\
}
\date{}

\begin{document}
\maketitle
\begin{abstract}
Sentence-level Quality Estimation (QE) of machine translation is traditionally formulated as a regression task, and the performance of QE models is typically measured by Pearson correlation with human labels.
Recent QE models have achieved previously-unseen levels of correlation with human judgments, but they rely on large multilingual contextualized language models that are computationally expensive and thus infeasible for many real-world applications.
In this work, we evaluate several model compression techniques for QE and find that, despite their popularity in other NLP tasks, they lead to poor performance in this regression setting.
We observe that a full model parameterization is required to achieve SoTA results in a regression task. 
However, we argue that the level of expressiveness of a model in a continuous range is unnecessary given the downstream applications of QE, 
and show that reframing QE as a classification problem and evaluating QE models using classification metrics would better reflect their actual performance in real-world applications. 
\todo{modify this claim}

\end{abstract}

\section{Introduction}
Quality Estimation (QE) \cite{specia-etal-2020-findings-wmt} is the task of predicting the quality of an automatically translated sentence at test time, without the need to rely on reference translations.
Formally, given a source sentence, $s$ and a translated sentence, $t$, the goal of QE is to learn a regression model, $m$ such that $m(s,t)$ returns a score that represents the quality of the translated sentence. 

There are many important applications of quality estimation, for example, translation companies use QE systems to identify mistranslated sentences, which would reduce post-editing costs and efforts, while online platforms use QE systems as filters to hide poorly translated sentences from end-users. Additionally, with the proliferation of mined parallel data obtained from web-crawls as a source of NMT training data~\cite{el2020searching,el2020massive}, QE has become an important tool in performing quality control on translations from models trained on noisy training data.

The performance of a QE system is usually measured by the correlation 
between predicted QE and human-annotated QE scores. 
However, the predictions of QE models are primarily used to make binary decisions \cite{zhou2020practical}:
only translations above a certain QE threshold would be given to a human for post-edition in a translation company, or shown to the user in an online platform.
Therefore, Pearson correlation might not be the best metric to evaluate the actual performance of the QE models in real-world use cases.

In recent iterations of the QE shared task at the Conference on Machine Translation (WMT) \cite{fonseca2019findings, specia-etal-2020-findings-wmt}, the top-performing QE systems have been built on large multilingual contextualized language models that were pre-trained on huge amounts of multilingual text data.
Further, these QE models are multilingual and work well in zero-shot scenarios \cite{sun-etal-2020-exploratory}.  
This characteristic makes them very appealing for real-life scenarios because it removes the need to train one bilingual model for every pair of languages.

However, these neural QE models contain millions of parameters and as such their memory and disk footprints are very large. Moreover,
at inference time they are often 
more computationally 
expensive than the upstream neural machine translation (NMT) models, making them unsuitable for  deployment in applications with low inference latency requirements or on devices with disk or memory constraints. 
In this paper we explore applying compression techniques to these large QE models to yield more practical, lower-latency, models while retaining state-of-the-art (SoTA) performance.

Our {\bf main contributions and findings} are:
\begin{enumerate}
    \item We conduct a thorough study on the efficiency of SoTA neural QE models.
    \item We shed light on the performance of recent compression techniques on a multilingual regression task and show that these techniques are inadequate for regression.
    \item We empirically show that regression has a lower level of compression effectiveness than classification, on  publicly available multilingual QE datasets.
    \item We argue that the level of expressiveness of a regression model in a continuous range is unnecessary given the downstream applications of QE, and evaluating QE models using classification metrics would better reflect their actual performance in real-world applications. 
    \todo{modify this claim}
    \item We find that multilingual QE models are not as effective as bilingual QE models on both regression and binary classification, for models with higher degrees of compression. 
\end{enumerate}
\section{Related Work}

Early work on QE built models on manually crafted features extracted from the source and translated sentences, or confidence features directly from machine translation (MT) systems \cite{specia2009estimating}.
In contrast, SoTA models are usually trained in an end-to-end manner using neural networks \cite{specia2018findings, fonseca2019findings, specia-etal-2020-findings-wmt, tuan2021quality}, without the additional step of feature extraction.

The proliferation of many-to-many NMT~\cite{fan2021beyond,ko2021adapting} has motivated similar multilingual QE models. These multilingual QE models have exploited large pre-trained contextualized multilingual language models to achieve a previously-unseen level of correlation with human judgments in recent iterations of the WMT QE shared task.  
For example, the top-performing QE model at WMT 2019 \cite{kepler2019unbabels} is a neural predictor-estimator model based on multilingual BERT  \cite{devlin-etal-2019-bert}, while
the best QE models at WMT 2020 \cite{transquest:2020,fomicheva-etal-2020-bergamot} are regression models built on XLM-R \cite{conneau2019unsupervised}.
\citet{sun-etal-2020-exploratory} find that these models generalize well across languages and training a single multilingual QE model is more effective than training a bilingual model for every language direction.
Unfortunately, these models are computationally infeasible for real-world applications.

Several approaches have been proposed to address the latency issues of these large contextualized language models.
Most of these work are based on \textbf{knowledge distillation} \cite{sanh2019distilbert, jiao2019tinybert, aguilar2020knowledge, tang2019distilling, sun2019patient} where large contextualized language models (teacher) are used to supervise the training of smaller student models \cite{hinton2015distilling}, or \textbf{pruning}, which discards redundant model components. 
Some examples are model weights \cite{gordon2020compressing}, tokens \cite{goyal2020power}, encoder layers \cite{sajjad2020poor} and attention heads \cite{michel2019sixteen}.

Existing work has looked at model compression for other multilingual tasks:
\citet{tsai-etal-2019-small} obtained 27x speedup on multilingual sequence labeling without any significant performance degradation, while \citet{mukherjee2020xtremedistil} obtain 51x speedups on NER tasks while retaining 95\% performance.
In our experiments with QE we do not observe the same levels of compression effectiveness, suggesting that the QE is a much harder task for model compression.

A call to reframe QE as a classification problem was made by \citet{zhou2020practical}, based on the perspective that classification is more suitable for real-world use cases and binary classes are easier to interpret than the predicted QE scores.
Our work suggests the same direction, but now from the perspective of modeling, where we empirically find that the level of expressiveness of a regression-based QE model is unnecessary and evaluating QE models using classification metrics would better reflect their actual performance in real-world applications.


\section{Background and hypothesis}

\begin{figure}[ht!]
\includegraphics[width=0.9\linewidth]{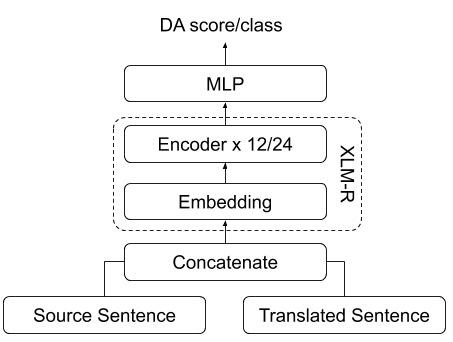}
\centering
\caption{XLM-R based neural QE model. The base version of XLM-R has 12 encoder layers while the large version of XLM-R has 24 encoder layers.}
\label{fig:xlmr}
\end{figure}

Current 
state of the art QE systems 
\cite{fomicheva-etal-2020-bergamot, transquest:2020, sun-etal-2020-exploratory}.
are built on XLM-R \cite{conneau2019unsupervised}, a contextualized language model pre-trained on more than 2 terabytes of filtered CommonCrawl data \cite{wenzek2020ccnet}.
As seen in Figure \ref{fig:xlmr}, the model concatenates a pair of source and translated sentences with a separator token in between and appends a special \textbf{CLS} token to the beginning of the concatenated string. 
It then converts the pre-processed string into a sequence of embedding vectors using a pre-trained embedding lookup table.
The embedding vectors are then encoded using the multi-head self-attention mechanism as described in \citet{vaswani2017attention}.
This step is repeated 12 times for the \xbase{} version of XLM-R and 24 times for the \xlarge{} version of XLM-R.
Finally, the neural QE model converts the final encoding of the \textbf{CLS} token into a QE score via a multilayer perceptron (MLP) layer.\\

In this work, we follow the neural architecture in \citet{sun-etal-2020-exploratory}, and experiment with both \textbf{bilingual models (BL)}, where each model is trained on data from only one language direction, and \textbf{multilingual models (ML)} that are trained on the concatenated data from all language pairs available in the dataset.
We choose mean squared error and binary cross-entropy as the objective functions for the regression and binary classification tasks respectively.
We use AdamW \cite{loshchilov2017decoupled} for parameter tuning and use a common learning rate of $1e^{-6}$ for all experiments.

\subsection{Efficiency of neural QE models}
\addtolength{\tabcolsep}{-1pt}
\begin{table}[h]
\centering
\begin{tabular}{ccccc}
\toprule
&\multicolumn{2}{c}{\textbf{\xbase{}}}&\multicolumn{2}{c}{\textbf{\xlarge{}}}\\
\textbf{Module}&\textbf{\#P}&\textbf{L(ms)}&\textbf{\#P}&\textbf{L(ms)}\\
\midrule
Embedding & 192M & 0.7 & 257M & 0.9 \\
Encoder & 7.1M & 10.0 & 12.6M & 15.3 \\
MLP & 2.4M & 0.6 & 4.2M & 0.8 \\
\midrule
Total & 280M & 122 & 563M & 370 \\
\bottomrule

\end{tabular}
\caption{
Number of parameters (\#P) and latency measured in milliseconds (L) for different components of the base and large version of XLM-R.
For the encoder, we show the average statistics for each layer. 
Note that the \xbase{} version has 12 encoder layers, while the large version has 24.
}
\label{benchmark}
\end{table}

To gain insights on the practicality of deploying the aforementioned baseline QE models in real-world situations, we gathered benchmark results on a server with 40 physical cores and 512GB of RAM. 
We measure the average time required to compute DA score for every sentence pair in the test sets across all seven language directions.
We use a batch size of 1 and run the QE models on CPU.
We highlight some of the findings in Table \ref{benchmark}.

\paragraph{Memory} 
XLM-R \xbase{} has around 280 million parameters, and 69\% of the parameters are in the embedding layer.
XLM-R \xlarge{} has around 563 million parameters, which is more than 2 times the number of parameters in XLM-R \xbase{}, and 54\% of the parameters are in the encoder layers.
Given that these models take up around 1-2 GB of memories on disk, they might be unsuitable for devices with small RAM capacity. 
\paragraph{Latency} 
Most of the computations take place in the encoder layers, where the latency of each encoder layer is 15.3 milliseconds for XLM-R \xlarge{} and 10 milliseconds for XLM-R \xbase{}.
Although the embedding layer contains a significant number of parameters, it requires much fewer computations than the encoder layers since its embedding matrices are only used as lookup tables.

Given the benchmark results in Table \ref{benchmark}, it is clear that recent QE models based on XLM-R \xlarge{} are computationally expensive and memory intensive. 
At an average inference time of 370 milliseconds per sentence pair, these QE models can be even slower than the upstream MT systems, making them infeasible in real-world applications that require real-time response. 
As more than 98\% of inference time is spent in the encoder layers of the neural QE models, we will explore model compression techniques that could reduce the number of parameters and computations in those layers.

\section{Model compression techniques}

Given the vast amount of work in the field of model compression, we explore three broad techniques and examine whether they could be successfully applied to compressing QE models.

\subsection{Pruning}
Pruning techniques are inspired by observations that large pre-trained contextualized language models might be over-parameterized for downstream tasks and most model parameters can be removed without significantly hurting model performance \cite{kovaleva2019revealing}. 

\paragraph{Layer pruning} \citet{sajjad2020poor} demonstrated that it is possible to drop encoder layers from pre-trained contextualized language models while maintaining up to 98\% of their original performance.
We apply the \emph{top-layer strategy} to XLM-R by dropping the top N encoders layers and then fine-tune the reduced neural architecture on QE datasets.  
We experiment with different values for N from \{3, 6, 9, 12, 15, 18, 21, 23\}.

\paragraph{Token pruning}
\citet{pmlr-v119-goyal20a} observed that token vectors start carrying similar information as they pass through the encoder layers of contextualized language models.
The authors propose a method that progressively removes redundant word vectors by only keeping the top K vectors at each encoder layer based on an attention scoring mechanism.
To determine the optimal value of K for each encoder layer, they  add a soft extraction layer with learnable parameters in the range [0, 1] that represents the degree of usefulness for every token vector.
L1 regularizers are used to optimize the weights of the parameters in the soft extraction layer.
Following the original implementation, we tune a hyper-parameter that controls the trade-off between the loss of the original tasks and the regularizers.

\subsection{Knowledge distillation}

Recent knowledge distillation (KD) methods \cite{jiao2019tinybert, mao2020ladabert, sanh2019distilbert} use larger BERT models (teacher) to supervise the training of smaller BERT models (student), typically with the help of the same raw text data that was used by the teacher models.
Given that XLM-R was trained on more than 2 terabytes of multilingual text data, it would be computationally difficult to adapt the KD techniques to XLM-R.
Instead, we experiment with a simplified KD setup inspired by \citet{xu2020bert}.

\paragraph{Module replacement} 
\begin{figure}[ht!]
\begin{center}
\includegraphics[width=0.7\linewidth]{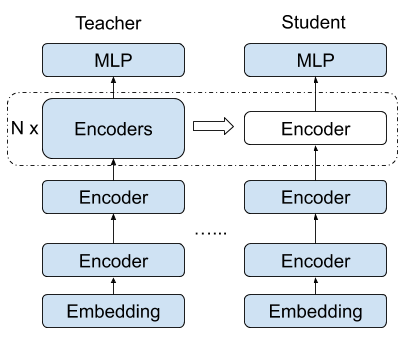}
\end{center}
\caption{
Module replacement: replacing N encoder layers with a single encoder layer.
}
\label{kd}
\end{figure}

We explore whether it is effective to compress N encoder layers into a single encoder layer.
As seen in Figure \ref{kd}, we use the top N layers of a fine-tuned QE model to supervise the training of one encoder layer in a smaller QE model.
For the student QE model, we randomly initialize the target encoder layer and copy all other parameters from the teacher QE model.
During training, we freeze all parameters except for the ones in the target encoder layer.
The loss function is defined as the sum of 1) the mean squared error between the output of the final teacher encoder and the output of the target student encoder, and 2) the original objective function.
We use sentence pairs in the MLQE dataset to train the student model and experiment with the following values for N: \{2, 6, 12, 18, 23, 24\}.
\section{Experimental settings}
We report results on the MLQE-PE dataset using  Pearson correlation for regression and F1 for classification.

\subsection{QE dataset}\label{sec:dataset}

MLQE-PE is the publicly released multilingual QE dataset used for the WMT 2020 shared task on QE  \cite{fomicheva2020mlqepe}. 
This dataset was built using source sentences extracted from Wikipedia and Reddit, translated to and from English with SoTA bilingual NMT systems that were trained on publicly available parallel corpora.
It contains seven language pairs:
the high-resource English-German (En-De), English-Chinese (En-Zh), and Russian-English (Ru-En); 
the medium-resource Romanian–English (Ro-En) and Estonian–English (Et-En); 
and the low-resource Sinhala–English (Si-En) and Nepali–English (Ne-En).
Each pair of sentences was manually annotated for quality using a  0--100 direct assessment (DA) scheme as shown in table \ref{annotation}. A z-normalized version of these scores is used directly for {\bf regression}.

\begin{table}[h]
\centering
\begin{tabular}{p{0.2\linewidth}p{0.7\linewidth}}
\toprule
\textbf{DA score}&\textbf{Meaning}\\
\midrule
0-10&incorrect translation\\
\midrule
11-29&translation with few correct keywords, but the overall meaning is different from the source\\
\midrule
30-50&a translation with major mistakes\\
\midrule
51-69&a translation which is understandable and conveys the overall meaning of the source but contains typos or grammatical errors\\
\midrule
70-90&a translation that closely preserves the semantics of the source sentence\\
\midrule
90-100&a perfect translation\\
\bottomrule

\end{tabular}
\caption{
Annotation scheme used to build the MLQE-PE dataset
}
\label{annotation}
\end{table}


As previously mentioned, since the most common use case of the QE is to make binary decisions based on predicted QE scores \cite{zhou2020practical}, i.e, to determine whether a translated sentence is adequate based on a certain threshold on QE score, we also experiment with models that are directly optimized for {\bf binary classification}.
To that end, we modify the MLQE-PE dataset to predict the \emph{acceptability} of a translation by assigning the label \emph{not acceptable} to sentence pairs with DA scores less than some threshold values and the label \emph{acceptable} for the remainder of the translations. 
The notion of acceptability is thus based on the guidelines provided for the MLQE-PE annotations in Table~ \ref{annotation}. 
Here, we require that a translation is understandable but not necessarily perfect, and experiment with two thresholds of $\ge51$ and $\ge70$ to signify acceptability.\footnote{We acknowledge that this threshold may be application-dependent. In other cases, where a higher level of quality is desired (e.g. for \emph{knowledge dissemination}), a $\ge90$ threshold might be more appropriate.}

\subsection{Evaluation metrics}
Following the standard practice used by the QE research community, we measure the performance of a QE model by calculating the \emph{Pearson Correlation coefficient} of its predicted DA scores and the actual human-annotated DA scores on a test set.
Formally, let $x_i=m(s_i, t_i)$ be the DA score for a sentence pair $(s_i, t_i)$ predicted by a QE model, m, and $y_i$ be the actual human-annotated DA score. 
Then the performance of m on a test set $T=\{(s_1, t_1, y_1), (s_2, t_2, y_2), \dots (s_N, t_N, y_N)\}$ is defined as:
\begin{equation}
    r=\dfrac{\sum\limits_{i=1}^N (x_i - \bar{x})(y_i - \bar{y})}{\sqrt{\sum\limits_{i=1}^N (x_i - \bar{x})^2 \sum\limits_{i=1}^N (y_i - \bar{y})^2}}  
\end{equation}
r is in the range $[-1, 1]$ and a better model would have a r closer to 1.

For binary classification, we use the F1 score defined as:
\begin{equation}
    F1 = \dfrac{2 \times (P \times R)}{(P + R)}
\end{equation}
where P is the precision and R is the recall.
For head-to-head comparison, we also evaluate regression models using F1 by converting the predicted QE scores into binary classes using the same threshold described in previous subsection.
\section{Results}

The baseline results on the MLQE dataset\footnote{\url{https://github.com/sheffieldnlp/mlqe-pe}} used at WMT 2020 are shown in Table \ref{baseline}.
We keep the same train-dev-test splits as the WMT 2020  shared task\footnote{\url{http://www.statmt.org/wmt20/quality-estimation-task.html}} for all experiments.
The results of our regression models are comparable to the results reported in recent work \cite{fomicheva-etal-2020-bergamot, ranasinghe2020transquest}.
In general, QE models based on XLM-R \xlarge{} outperform models based on XLM-R \xbase{}, showing that a higher number of parameters benefits the QE task.
This is especially true for the regression tasks, where on average, the large models outperform the base models by 11\% and 56.8\% for bilingual and multilingual models respectively.
However, the same levels of performance degradation are not observed on the classification tasks: On average, the large QE models only perform 3.7\%/5.6\% and 1.2\%/7.1\% better than the base QE models for bilingual and multilingual settings respectively at different thresholds.
This shows that classification performance depends less on the number of model parameters and therefore we could potentially observe better compression results and more accurate model performance in real-world application if we evaluate QE with classification metrics.
\todo{modify this claim}
In the remaining of this section we explore different compression techniques on both regression and classification to test this hypothesis.

\addtolength{\tabcolsep}{-3pt}
\begin{table}[ht!]
\small
\centering
\begin{tabular}{cccccc}

\toprule
\multirow{2}{*}{\textbf{Model}} & \multirow{2}{*}{\textbf{Type}} & \multicolumn{2}{c}{\begin{tabular}{c}\textbf{Regression}\\ corr. ($\rho$)\end{tabular}}& \multicolumn{2}{c}{\begin{tabular}{c}\textbf{Classification}\\ F1 ($\geq51$ / $\geq70$)\end{tabular}} \\
\cmidrule(r){3-4}
\cmidrule(l){5-6}
 &  & \textbf{Base} & \textbf{Large} & \textbf{Base} & \textbf{Large} \\
\midrule
BERGAMOT&BL&-&0.67&-&-\\
-LATTE&ML&-&0.69&-&-\\
\midrule
\multirow{2}{*}{TransQuest}&BL&-&0.69&-&-\\
&ML&-&0.67&-&-\\
\midrule
\multirow{2}{*}{This work}&BL&0.63&0.70&0.82/0.72&0.85/0.76\\
&ML&0.44&0.69&0.83/0.70&0.84/0.75\\

\bottomrule
\end{tabular}
\caption{
Baseline results of bilingual (BL) and multilingual (ML) models on the MLQE-PE dataset for both regression and binary classification at different thresholds. 
Results are averaged over 5 different runs and 7 language directions.
Our results are comparable to the results of BERGAMOT-LATTE \cite{fomicheva-etal-2020-bergamot} and TransQuest \cite{ranasinghe2020transquest}, the top-performing systems at WMT 2020 QE shared task.
}
\label{baseline}
\end{table}

\subsection{Compression techniques results}
\begin{figure}[ht!]
\centering
\includegraphics[width=0.95\linewidth]{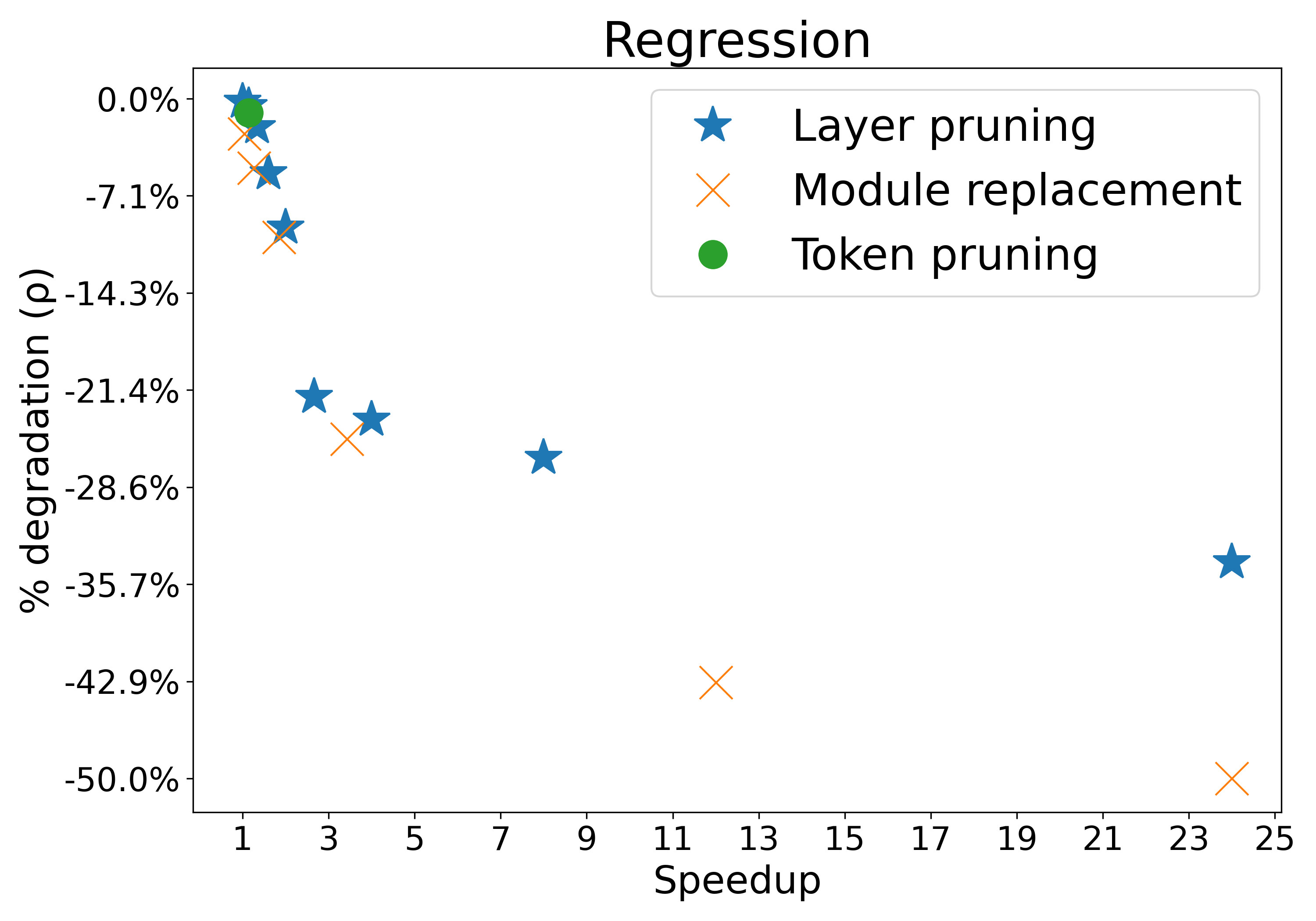}
\caption{
\% degradation ($\rho$) vs. speedup for regression models.
}
\label{reg_speedup}
\end{figure}

\begin{figure}[ht!]
\centering
\includegraphics[width=0.95\linewidth]{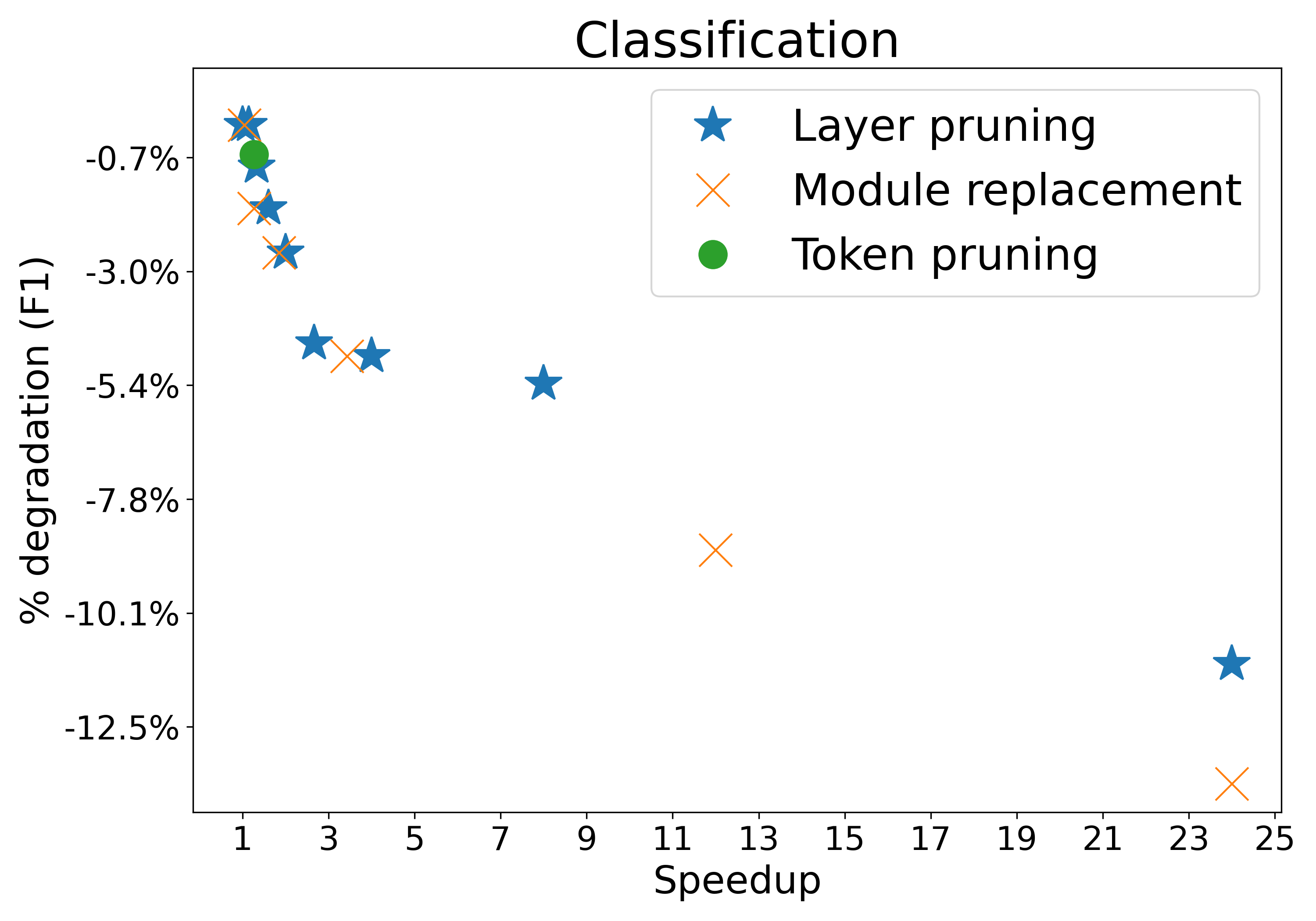}
\caption{
\% degradation ($F1$) vs. speedup for classification models ($\geq$ 51).
}
\label{cls_speedup}
\end{figure}

We apply each compression technique to bilingual QE models using XLM-R \xlarge{}, and average the results over 5 different runs for every language direction.
We then compute the average speedup of every compressed model to its original model.
The performance drop against speedup plots\footnote{averaged over 7 language directions} of regression models and classification models are shown in Figure \ref{reg_speedup} and Figure \ref{cls_speedup} respectively.

Based on the results of our experiments, we find that layer pruning outperforms module replacement, especially at higher speedup.
For example, at around 24x speedup, layer pruning outperforms module replacement by 31.8\% and 2.9\% for regression and classification respectively.
For token pruning, we find that it does not offer any significant benefit over the other 2 compression techniques.
Further, the token-pruned models that we tuned to the best of our efforts are conservative and prefer model accuracy over speedup: 
14.4\% and 26.4\% faster than the original QE models with $<$ 1\% performance degradation for regression and classification respectively.
In the remainder of this paper, we will focus on running model compression experiments with the layer pruning strategy.

\subsection{Compression techniques are inadequate when evaluating QE as a regression task}
As seen in Figure \ref{reg_speedup}, the compressed regression-based QE models have a performance degradation of more than 9\% at just 2x speedup.
The performance degradation worsens to 23\% at 4x speedup and 34\% at 24x speedup.
These results are significantly worse than the numbers reported on other NLP tasks: \citet{jiao2019tinybert} reported a 9.4x speedup with 96.8\% performance retention on GLUE benchmark \cite{wang2018glue} , \citet{mukherjee2020xtremedistil} reported 51x speedup with 95\% performance retention on a multilingual NER task with 41 languages and \cite{wang2020minilm} reported 2x speedup with more than 99\% performance retention on SQUAD 2.0 \cite{rajpurkar2018know}.

Our results suggest that QE datasets might not  suffer from the same degree of \emph{overparameterization problem} \citet{kovaleva2019revealing} observed in other NLP datasets.
We hypothesize that the performance of QE \emph{depends heavily on the large number of parameters in XLM-R}.
This is further supported by the results in Table \ref{baseline}, where the XLM-R \xlarge{} models, with around twice the number of parameters in XLM-R base models, outperform the latter by more than 11\%.

\subsection{Better compression results when evaluating QE with classification metric}
\todo{new subsection title}
\begin{table}[ht!]
\small
\centering
\begin{tabular}{ccccc}
\toprule
&\textbf{En-De}&\textbf{En-Zh}&\textbf{Ru-En}&\textbf{Ro-En}\\
\midrule
Regr. & -8.0\%&-13.7\%&-8.8\%&-4.5\%\\
Class. ($\geq51$) & 0.0\%&0.0\%&-1.1\%&-1.1\%\\
Class. ($\geq70$)& 0.0\%&0.0\%&-2.8\%&-3.9\%\\
\midrule
& \textbf{Et-En}&\textbf{Si-En}&\textbf{Ne-En}&\textbf{Average}\\
\midrule
Regr. & -14.1\%&-9.2\%&-10.3\%&-9.8\%\\
Class. ($\geq51$)& -6.0\%&-1.4\%&-12.7\%&-3.2\%\\
Class. ($\geq70$)& -12.1\%&-0.1\%&-9.1\%&-4.0\%\\
\bottomrule
\end{tabular}
\caption{
Performance drops of \xbase{} QE models with respect to \xlarge{} QE models for regression (top) and binary classification at different thresholds.
}
\label{gain}
\end{table}

In Table \ref{gain}, we compute the relative percentage of performance drops when using XLM-R \xbase{} instead of XLM-R \xlarge{}.
We observe that the base QE models perform significantly worse than the large QE models on regression tasks, with an average performance drop of 9.8\% over 7 language directions.
In contrast, the average performance drops of 3.2\% and 4.0\% for binary classification at different thresholds are significantly lower, with less than 1.5\% drop on majority of 7 language directions.
Figure \ref{cls_speedup} also shows that it is possible to retain 94.6\% performance with 8x speedup and 88.8\% with 24x speedup.

Comparing these to the results in the previous subsection, we observe significantly less model degradation when evaluating the QE models with classification metric instead of regression metric.
\todo{tweak this claim}

\subsection{Regression or binary classification?}

\begin{figure}[ht!]
\centering
\includegraphics[width=0.98\linewidth]{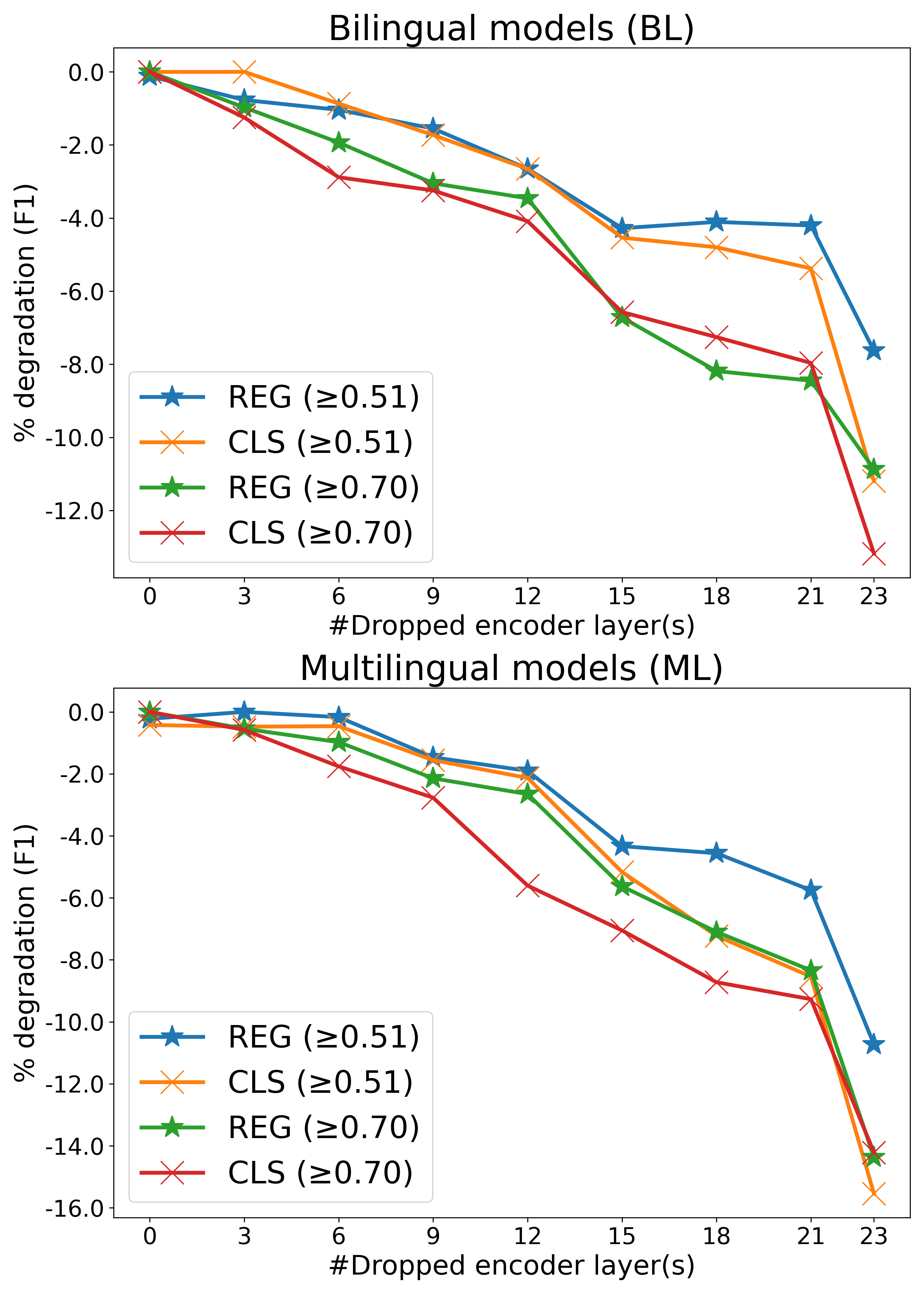}
\caption{
Comparison between regression models (REG) and classification models (CLS) for different layer pruning configurations.
All models are based on XLM-R \xlarge{}.
Top shows the results for bilingual models and bottom shows the results for multilingual models.
}
\label{reg_or_cls}
\end{figure}

In practice, the predicted QE scores from regression models are primarily used to make binary decisions based on predetermined thresholds.
To test whether it would be better to directly optimize QE models for binary classification, we convert the predicted DA scores from regression models into binary classes using the same threshold as the one in Section \ref{sec:dataset} and then compute F1 scores.

As shown in Figure \ref{reg_or_cls}, the F1 performances of regression models are comparable to the classification models at lower compression settings.
However, the regression models start to outperform the classification models when we further compress the models by dropping more encoder layers.
Our results show that in our case of using thresholds of $\geq51$ and $\geq70$ with layer pruning as our model compression strategy, optimizing QE as a regression task and then converting the predicted DA scores into binary classification labels seems better than directly optimizing QE as a binary classification task.

\subsection{Pearson correlation is misleading}
\begin{figure}[ht!]
\centering
\includegraphics[width=0.95\linewidth]{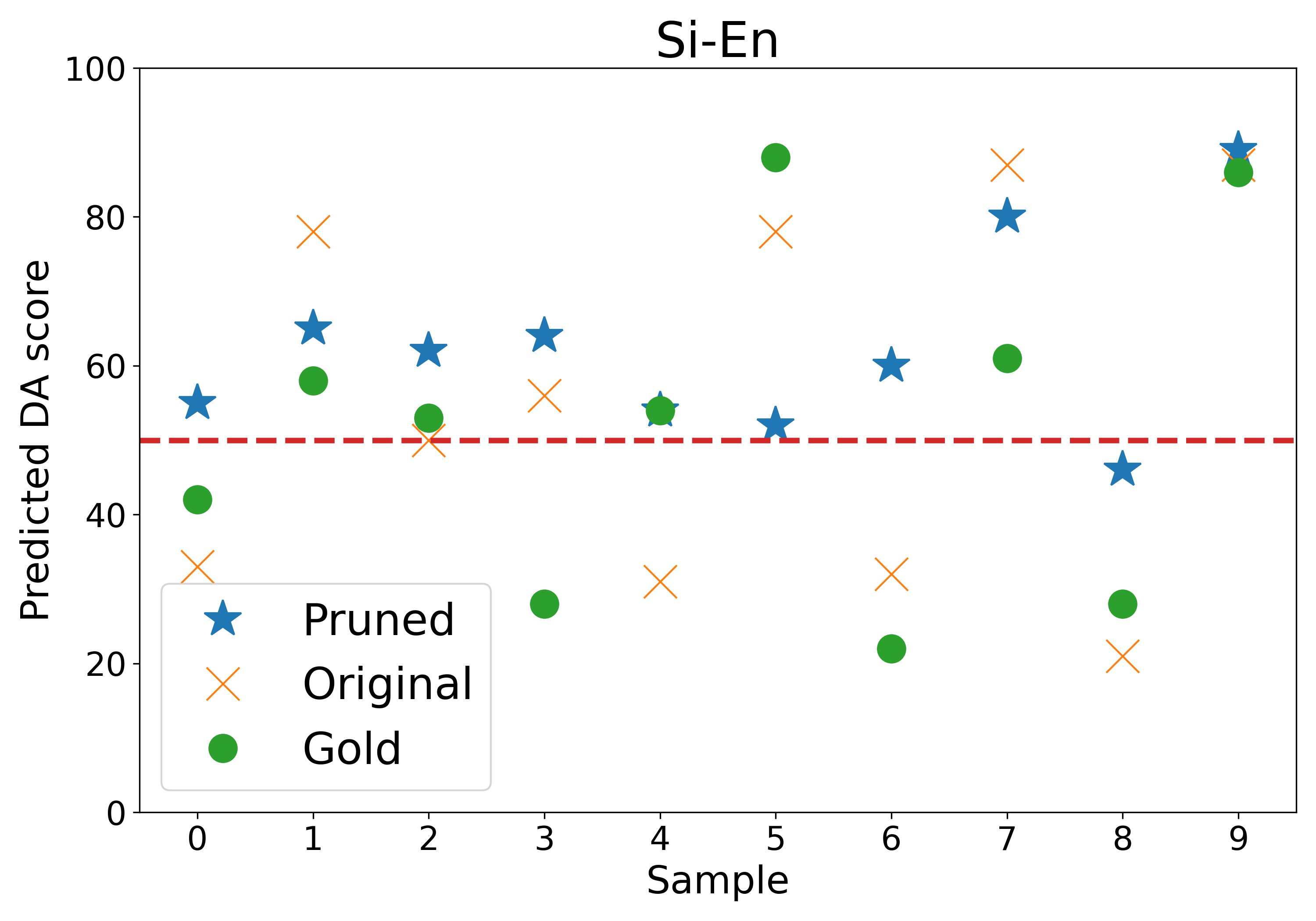}
\caption{
Predicted DA scores from a pruned and an original QE model, as well as the gold labels for 10 random samples from Sinhala-English test set.
Pearson correlation is 0.47 and 0.64, and F1 is 0.72 and 0.73 for the pruned and original models respectively.
The red dashed line is the decision boundary.
}
\label{pc_points}
\end{figure}

Looking at the results in Figures \ref{reg_speedup}, \ref{cls_speedup} and \ref{reg_or_cls}, it is apparent that the drastic drops in Pearson correlation at higher compression settings do not translate to equivalent degrees of performance degradation in terms of F1.
For example, at approximately 24x speedup, the bilingual regression models suffer an average performance degradation of 33.9\% in Pearson correlation, which is significantly higher than the 7.5\% performance degradation observed in binary classification.

The explanation lies in the fact that Pearson correlation penalizes predictions that do not follow the same linear trend as the gold DA scores. 
However, getting the linear trend right is not useful for binary classification, i.e., a predicted DA score is correct as long as it falls on the right side of the decision boundary, regardless of the degree of closeness to the gold DA score.
This phenomenon is illustrated in Figure \ref{pc_points} where the predicted scores of a QE model based on XLM-R \xlarge{} are generally closer to the gold DA scores than the predicted scores of a layer pruned QE model, which explains their 36.2\% difference in Pearson correlation.
However, the two models obtain comparable F1 scores because their predicted DA scores generally fall on the same sides of the decision boundary.

Our results suggest that larger QE models are required to make more accurate QE score predictions that have a higher Pearson correlation with human-annotated QE scores.
However, the higher accuracy and more optimal ordering of the predicted QE scores do not necessarily contribute to higher accuracy when making binary judgments.
We believe that Pearson correlation is a misleading evaluation metric that deviates from the use cases in real-world settings.
Chasing higher Pearson correlation could lead us down the path of building larger models that are computationally infeasible, yet having better exact DA predictions is not necessarily useful for better binary classification.

Based on these results, we recommend that \emph{classification metrics are used to evaluate the effectiveness of compressed QE models.}

\subsection{Does model compression affect the performance of multilingual QE?}

\begin{figure}[h]
\centering
\includegraphics[width=0.95\linewidth]{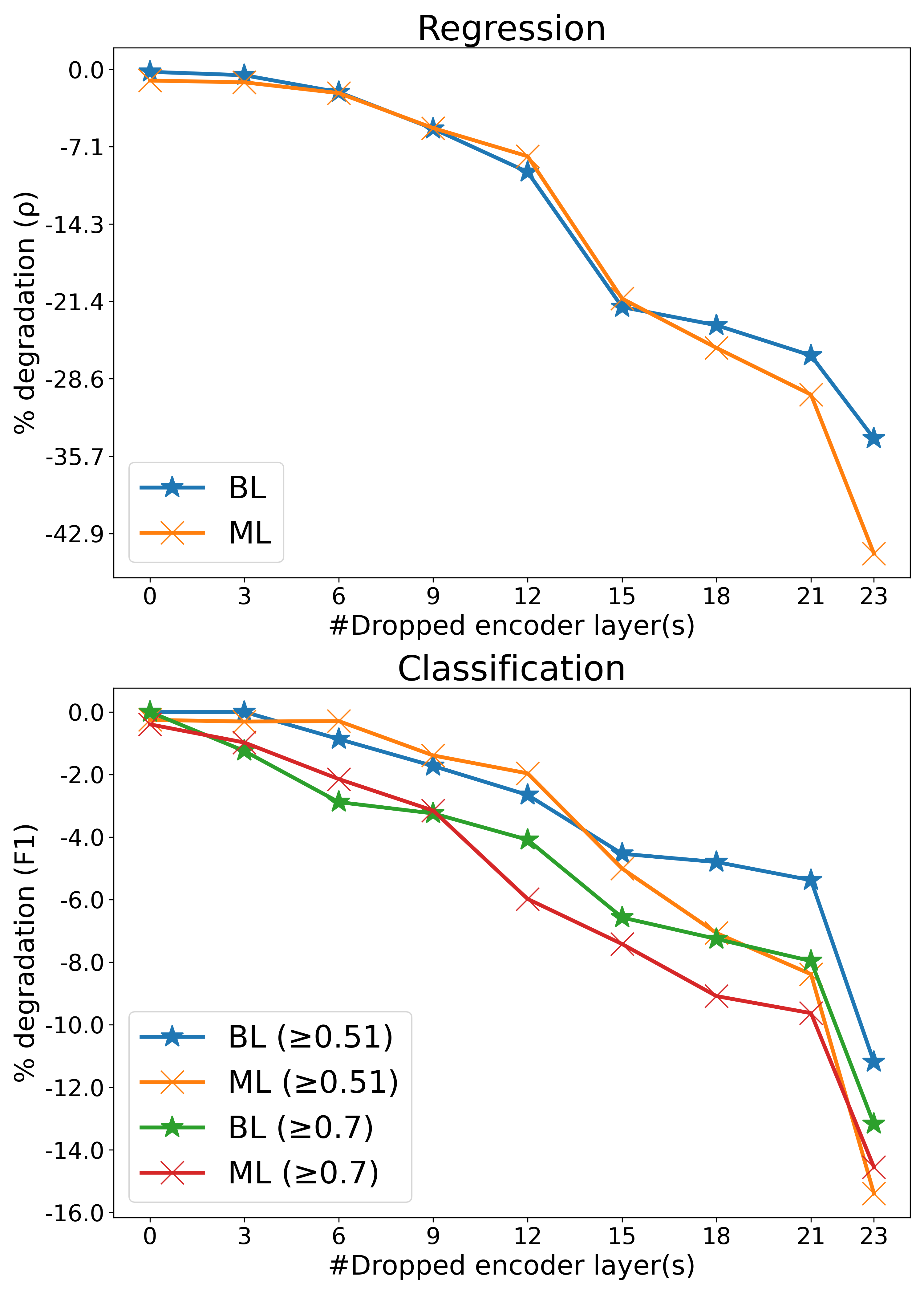}
\caption{Comparison of multilingual models (ML) against bilingual models (BL) for different layer pruning configurations. }
\label{multilingual_power}
\end{figure}

We plot the averaged results of bilingual and multilingual models for different configurations of layer pruning in Figure \ref{multilingual_power}.
We observe that at lower compression settings with less than or equals to 12 dropped layers, the results of the multilingual models are generally comparable to the results of the bilingual models and this corroborates the finding that multilingual QE model generalize well across languages \cite{sun-etal-2020-exploratory}.
However, at higher compression settings with more than 12 dropped layers, the multilingual models start to lose their effectiveness, underperforming the bilingual models significantly.
We hypothesize that the multilingual neural models no longer have enough model capacities for 7 different language directions beyond a certain degree of model compression.

\section{Conclusions}
This paper presents a thorough study on the efficiency of SoTA neural QE models and explores whether recent compression techniques can be successfully applied to reduce the size and improve the latency of these models.
Our experimental results show that recent compression techniques are inadequate for regression as we observe significant performance degradation on the QE task with little improvement in model efficiency.
We argue that the level of expressiveness of a QE model in a continuous range is unnecessary since the outputs of the QE model are usually used to make binary decisions.
Our results show that it is more appropriate to reframe the QE task as a classification problem, and evaluating QE models using classification metrics would better reflect their actual performance in real-world applications.
This enables us to achieve SoTA performance with tiny and efficient models.

Our experimental results suggest that compressing large neural QE models for the QE regression task remains a challenging problem, especially in the case of multilingual models, where they start showing higher degrees of performance degradation than their bilingual counterparts.


\bibliography{anthology,custom}

\begin{thebibliography}{40}
\expandafter\ifx\csname natexlab\endcsname\relax\def\natexlab#1{#1}\fi

\bibitem[{Aguilar et~al.(2020)Aguilar, Ling, Zhang, Yao, Fan, and
  Guo}]{aguilar2020knowledge}
Gustavo Aguilar, Yuan Ling, Yu~Zhang, Benjamin Yao, Xing Fan, and Chenlei Guo.
  2020.
\newblock Knowledge distillation from internal representations.
\newblock In \emph{Proceedings of the AAAI Conference on Artificial
  Intelligence}, volume~34, pages 7350--7357.

\bibitem[{Conneau et~al.(2019)Conneau, Khandelwal, Goyal, Chaudhary, Wenzek,
  Guzm{\'a}n, Grave, Ott, Zettlemoyer, and Stoyanov}]{conneau2019unsupervised}
Alexis Conneau, Kartikay Khandelwal, Naman Goyal, Vishrav Chaudhary, Guillaume
  Wenzek, Francisco Guzm{\'a}n, Edouard Grave, Myle Ott, Luke Zettlemoyer, and
  Veselin Stoyanov. 2019.
\newblock Unsupervised cross-lingual representation learning at scale.
\newblock \emph{arXiv preprint arXiv:1911.02116}.

\bibitem[{Devlin et~al.(2019)Devlin, Chang, Lee, and
  Toutanova}]{devlin-etal-2019-bert}
Jacob Devlin, Ming-Wei Chang, Kenton Lee, and Kristina Toutanova. 2019.
\newblock \href {https://doi.org/10.18653/v1/N19-1423} {{BERT}: Pre-training of
  deep bidirectional transformers for language understanding}.
\newblock In \emph{Proceedings of the 2019 Conference of the North {A}merican
  Chapter of the Association for Computational Linguistics: Human Language
  Technologies, Volume 1 (Long and Short Papers)}, pages 4171--4186,
  Minneapolis, Minnesota. Association for Computational Linguistics.

\bibitem[{El-Kishky et~al.(2020{\natexlab{a}})El-Kishky, Chaudhary, Guzm{\'a}n,
  and Koehn}]{el2020massive}
Ahmed El-Kishky, Vishrav Chaudhary, Francisco Guzm{\'a}n, and Philipp Koehn.
  2020{\natexlab{a}}.
\newblock A massive collection of cross-lingual web-document pairs.
\newblock In \emph{Proceedings of the 2020 Conference on Empirical Methods in
  Natural Language Processing (EMNLP)}, pages 5960--5969.

\bibitem[{El-Kishky et~al.(2020{\natexlab{b}})El-Kishky, Koehn, and
  Schwenk}]{el2020searching}
Ahmed El-Kishky, Philipp Koehn, and Holger Schwenk. 2020{\natexlab{b}}.
\newblock Searching the web for cross-lingual parallel data.
\newblock In \emph{Proceedings of the 43rd International ACM SIGIR Conference
  on Research and Development in Information Retrieval}, pages 2417--2420.

\bibitem[{Fan et~al.(2021)Fan, Bhosale, Schwenk, Ma, El-Kishky, Goyal, Baines,
  Celebi, Wenzek, Chaudhary et~al.}]{fan2021beyond}
Angela Fan, Shruti Bhosale, Holger Schwenk, Zhiyi Ma, Ahmed El-Kishky,
  Siddharth Goyal, Mandeep Baines, Onur Celebi, Guillaume Wenzek, Vishrav
  Chaudhary, et~al. 2021.
\newblock Beyond english-centric multilingual machine translation.
\newblock \emph{Journal of Machine Learning Research}, 22(107):1--48.

\bibitem[{Fomicheva et~al.(2020{\natexlab{a}})Fomicheva, Sun, Fonseca, Blain,
  Chaudhary, Guzmán, Lopatina, Specia, and Martins}]{fomicheva2020mlqepe}
Marina Fomicheva, Shuo Sun, Erick Fonseca, Frédéric Blain, Vishrav Chaudhary,
  Francisco Guzmán, Nina Lopatina, Lucia Specia, and André F.~T. Martins.
  2020{\natexlab{a}}.
\newblock \href {http://arxiv.org/abs/2010.04480} {Mlqe-pe: A multilingual
  quality estimation and post-editing dataset}.

\bibitem[{Fomicheva et~al.(2020{\natexlab{b}})Fomicheva, Sun, Yankovskaya,
  Blain, Chaudhary, Fishel, Guzm{\'a}n, and
  Specia}]{fomicheva-etal-2020-bergamot}
Marina Fomicheva, Shuo Sun, Lisa Yankovskaya, Fr{\'e}d{\'e}ric Blain, Vishrav
  Chaudhary, Mark Fishel, Francisco Guzm{\'a}n, and Lucia Specia.
  2020{\natexlab{b}}.
\newblock \href {https://www.aclweb.org/anthology/2020.wmt-1.116}
  {{BERGAMOT}-{LATTE} submissions for the {WMT}20 quality estimation shared
  task}.
\newblock In \emph{Proceedings of the Fifth Conference on Machine Translation},
  pages 1010--1017, Online. Association for Computational Linguistics.

\bibitem[{Fonseca et~al.(2019)Fonseca, Yankovskaya, Martins, Fishel, and
  Federmann}]{fonseca2019findings}
Erick Fonseca, Lisa Yankovskaya, Andr{\'e}~FT Martins, Mark Fishel, and
  Christian Federmann. 2019.
\newblock Findings of the wmt 2019 shared tasks on quality estimation.
\newblock In \emph{Proceedings of the Fourth Conference on Machine Translation
  (Volume 3: Shared Task Papers, Day 2)}, pages 1--10.

\bibitem[{Gordon et~al.(2020)Gordon, Duh, and Andrews}]{gordon2020compressing}
Mitchell Gordon, Kevin Duh, and Nicholas Andrews. 2020.
\newblock Compressing bert: Studying the effects of weight pruning on transfer
  learning.
\newblock In \emph{Proceedings of the 5th Workshop on Representation Learning
  for NLP}, pages 143--155.

\bibitem[{Goyal et~al.()Goyal, Choudhury, Raje, Chakaravarthy, Sabharwal, and
  Verma}]{pmlr-v119-goyal20a}
Saurabh Goyal, Anamitra~Roy Choudhury, Saurabh Raje, Venkatesan Chakaravarthy,
  Yogish Sabharwal, and Ashish Verma.
\newblock {P}o{WER}-{BERT}: Accelerating {BERT} inference via progressive
  word-vector elimination.

\bibitem[{Goyal et~al.(2020)Goyal, Choudhury, Raje, Chakaravarthy, Sabharwal,
  and Verma}]{goyal2020power}
Saurabh Goyal, Anamitra~Roy Choudhury, Saurabh Raje, Venkatesan Chakaravarthy,
  Yogish Sabharwal, and Ashish Verma. 2020.
\newblock Power-bert: Accelerating bert inference via progressive word-vector
  elimination.
\newblock In \emph{International Conference on Machine Learning}, pages
  3690--3699. PMLR.

\bibitem[{Hinton et~al.(2015)Hinton, Vinyals, and Dean}]{hinton2015distilling}
Geoffrey Hinton, Oriol Vinyals, and Jeff Dean. 2015.
\newblock Distilling the knowledge in a neural network.
\newblock \emph{arXiv preprint arXiv:1503.02531}.

\bibitem[{Jiao et~al.(2019)Jiao, Yin, Shang, Jiang, Chen, Li, Wang, and
  Liu}]{jiao2019tinybert}
Xiaoqi Jiao, Yichun Yin, Lifeng Shang, Xin Jiang, Xiao Chen, Linlin Li, Fang
  Wang, and Qun Liu. 2019.
\newblock Tinybert: Distilling bert for natural language understanding.
\newblock \emph{arXiv preprint arXiv:1909.10351}.

\bibitem[{Kepler et~al.(2019)Kepler, Trénous, Treviso, Vera, Góis, Farajian,
  Lopes, and Martins}]{kepler2019unbabels}
Fabio Kepler, Jonay Trénous, Marcos Treviso, Miguel Vera, António Góis,
  M.~Amin Farajian, António~V. Lopes, and André F.~T. Martins. 2019.
\newblock \href {http://arxiv.org/abs/1907.10352} {Unbabel's participation in
  the wmt19 translation quality estimation shared task}.

\bibitem[{Ko et~al.(2021)Ko, El-Kishky, Renduchintala, Chaudhary, Goyal,
  Guzm{\'a}n, Fung, Koehn, and Diab}]{ko2021adapting}
Wei-Jen Ko, Ahmed El-Kishky, Adithya Renduchintala, Vishrav Chaudhary, Naman
  Goyal, Francisco Guzm{\'a}n, Pascale Fung, Philipp Koehn, and Mona Diab.
  2021.
\newblock Adapting high-resource nmt models to translate low-resource related
  languages without parallel data.
\newblock \emph{arXiv preprint arXiv:2105.15071}.

\bibitem[{Kovaleva et~al.(2019)Kovaleva, Romanov, Rogers, and
  Rumshisky}]{kovaleva2019revealing}
Olga Kovaleva, Alexey Romanov, Anna Rogers, and Anna Rumshisky. 2019.
\newblock Revealing the dark secrets of bert.
\newblock In \emph{Proceedings of the 2019 Conference on Empirical Methods in
  Natural Language Processing and the 9th International Joint Conference on
  Natural Language Processing (EMNLP-IJCNLP)}, pages 4365--4374.

\bibitem[{Loshchilov and Hutter(2017)}]{loshchilov2017decoupled}
Ilya Loshchilov and Frank Hutter. 2017.
\newblock Decoupled weight decay regularization.
\newblock \emph{arXiv preprint arXiv:1711.05101}.

\bibitem[{Mao et~al.(2020)Mao, Wang, Wu, Zhang, Wang, Yang, Zhang, Tong, and
  Bai}]{mao2020ladabert}
Yihuan Mao, Yujing Wang, Chufan Wu, Chen Zhang, Yang Wang, Yaming Yang, Quanlu
  Zhang, Yunhai Tong, and Jing Bai. 2020.
\newblock Ladabert: Lightweight adaptation of bert through hybrid model
  compression.
\newblock \emph{arXiv preprint arXiv:2004.04124}.

\bibitem[{Michel et~al.(2019)Michel, Levy, and Neubig}]{michel2019sixteen}
Paul Michel, Omer Levy, and Graham Neubig. 2019.
\newblock Are sixteen heads really better than one?
\newblock \emph{arXiv preprint arXiv:1905.10650}.

\bibitem[{Mukherjee and Awadallah(2020)}]{mukherjee2020xtremedistil}
Subhabrata Mukherjee and Ahmed~Hassan Awadallah. 2020.
\newblock Xtremedistil: Multi-stage distillation for massive multilingual
  models.
\newblock In \emph{Proceedings of the 58th Annual Meeting of the Association
  for Computational Linguistics}, pages 2221--2234.

\bibitem[{Rajpurkar et~al.(2018)Rajpurkar, Jia, and Liang}]{rajpurkar2018know}
Pranav Rajpurkar, Robin Jia, and Percy Liang. 2018.
\newblock Know what you don’t know: Unanswerable questions for squad.
\newblock In \emph{Proceedings of the 56th Annual Meeting of the Association
  for Computational Linguistics (Volume 2: Short Papers)}, pages 784--789.

\bibitem[{Ranasinghe et~al.(2020{\natexlab{a}})Ranasinghe, Orasan, and
  Mitkov}]{transquest:2020}
Tharindu Ranasinghe, Constantin Orasan, and Ruslan Mitkov. 2020{\natexlab{a}}.
\newblock Transquest at wmt2020: Sentence-level direct assessment.
\newblock In \emph{Proceedings of the Fifth Conference on Machine Translation}.

\bibitem[{Ranasinghe et~al.(2020{\natexlab{b}})Ranasinghe, Orasan, and
  Mitkov}]{ranasinghe2020transquest}
Tharindu Ranasinghe, Constantin Orasan, and Ruslan Mitkov. 2020{\natexlab{b}}.
\newblock Transquest: Translation quality estimation with cross-lingual
  transformers.
\newblock In \emph{Proceedings of the 28th International Conference on
  Computational Linguistics}, pages 5070--5081.

\bibitem[{Sajjad et~al.(2020)Sajjad, Dalvi, Durrani, and
  Nakov}]{sajjad2020poor}
Hassan Sajjad, Fahim Dalvi, Nadir Durrani, and Preslav Nakov. 2020.
\newblock Poor man's bert: Smaller and faster transformer models.
\newblock \emph{arXiv preprint arXiv:2004.03844}.

\bibitem[{Sanh et~al.(2019)Sanh, Debut, Chaumond, and
  Wolf}]{sanh2019distilbert}
Victor Sanh, Lysandre Debut, Julien Chaumond, and Thomas Wolf. 2019.
\newblock Distilbert, a distilled version of bert: smaller, faster, cheaper and
  lighter.
\newblock \emph{arXiv preprint arXiv:1910.01108}.

\bibitem[{Specia et~al.(2020)Specia, Blain, Fomicheva, Fonseca, Chaudhary,
  Guzm{\'a}n, and Martins}]{specia-etal-2020-findings-wmt}
Lucia Specia, Fr{\'e}d{\'e}ric Blain, Marina Fomicheva, Erick Fonseca, Vishrav
  Chaudhary, Francisco Guzm{\'a}n, and Andr{\'e} F.~T. Martins. 2020.
\newblock \href {https://www.aclweb.org/anthology/2020.wmt-1.79} {Findings of
  the {WMT} 2020 shared task on quality estimation}.
\newblock In \emph{Proceedings of the Fifth Conference on Machine Translation},
  pages 743--764, Online. Association for Computational Linguistics.

\bibitem[{Specia et~al.(2018)Specia, Blain, Logacheva, Astudillo, and
  Martins}]{specia2018findings}
Lucia Specia, Fr{\'e}d{\'e}ric Blain, Varvara Logacheva, Ram{\'o}n Astudillo,
  and Andr{\'e}~FT Martins. 2018.
\newblock Findings of the wmt 2018 shared task on quality estimation.
\newblock In \emph{Proceedings of the Third Conference on Machine Translation:
  Shared Task Papers}, pages 689--709.

\bibitem[{Specia et~al.(2009)Specia, Turchi, Cancedda, Dymetman, and
  Cristianini}]{specia2009estimating}
Lucia Specia, Marco Turchi, Nicola Cancedda, Marc Dymetman, and Nello
  Cristianini. 2009.
\newblock Estimating the sentence-level quality of machine translation systems.
\newblock In \emph{13th Annual Conference of the European Association for
  Machine Translation}, pages 399--410.

\bibitem[{Sun et~al.(2020)Sun, Fomicheva, Blain, Chaudhary, El-Kishky,
  Renduchintala, Guzm{\'a}n, and Specia}]{sun-etal-2020-exploratory}
Shuo Sun, Marina Fomicheva, Fr{\'e}d{\'e}ric Blain, Vishrav Chaudhary, Ahmed
  El-Kishky, Adithya Renduchintala, Francisco Guzm{\'a}n, and Lucia Specia.
  2020.
\newblock \href {https://www.aclweb.org/anthology/2020.aacl-main.39} {An
  exploratory study on multilingual quality estimation}.
\newblock In \emph{Proceedings of the 1st Conference of the Asia-Pacific
  Chapter of the Association for Computational Linguistics and the 10th
  International Joint Conference on Natural Language Processing}, pages
  366--377, Suzhou, China. Association for Computational Linguistics.

\bibitem[{Sun et~al.(2019)Sun, Cheng, Gan, and Liu}]{sun2019patient}
Siqi Sun, Yu~Cheng, Zhe Gan, and Jingjing Liu. 2019.
\newblock Patient knowledge distillation for bert model compression.
\newblock In \emph{Proceedings of the 2019 Conference on Empirical Methods in
  Natural Language Processing and the 9th International Joint Conference on
  Natural Language Processing (EMNLP-IJCNLP)}, pages 4314--4323.

\bibitem[{Tang et~al.(2019)Tang, Lu, Liu, Mou, Vechtomova, and
  Lin}]{tang2019distilling}
Raphael Tang, Yao Lu, Linqing Liu, Lili Mou, Olga Vechtomova, and Jimmy Lin.
  2019.
\newblock Distilling task-specific knowledge from bert into simple neural
  networks.
\newblock \emph{arXiv preprint arXiv:1903.12136}.

\bibitem[{Tsai et~al.(2019)Tsai, Riesa, Johnson, Arivazhagan, Li, and
  Archer}]{tsai-etal-2019-small}
Henry Tsai, Jason Riesa, Melvin Johnson, Naveen Arivazhagan, Xin Li, and Amelia
  Archer. 2019.
\newblock \href {https://doi.org/10.18653/v1/D19-1374} {Small and practical
  {BERT} models for sequence labeling}.
\newblock In \emph{Proceedings of the 2019 Conference on Empirical Methods in
  Natural Language Processing and the 9th International Joint Conference on
  Natural Language Processing (EMNLP-IJCNLP)}, pages 3632--3636, Hong Kong,
  China. Association for Computational Linguistics.

\bibitem[{Tuan et~al.(2021)Tuan, El-Kishky, Renduchintala, Chaudhary,
  Guzm{\'a}n, and Specia}]{tuan2021quality}
Yi-Lin Tuan, Ahmed El-Kishky, Adithya Renduchintala, Vishrav Chaudhary,
  Francisco Guzm{\'a}n, and Lucia Specia. 2021.
\newblock Quality estimation without human-labeled data.
\newblock \emph{arXiv preprint arXiv:2102.04020}.

\bibitem[{Vaswani et~al.(2017)Vaswani, Shazeer, Parmar, Uszkoreit, Jones,
  Gomez, Kaiser, and Polosukhin}]{vaswani2017attention}
Ashish Vaswani, Noam Shazeer, Niki Parmar, Jakob Uszkoreit, Llion Jones,
  Aidan~N Gomez, {\L}ukasz Kaiser, and Illia Polosukhin. 2017.
\newblock Attention is all you need.
\newblock In \emph{Advances in neural information processing systems}, pages
  5998--6008.

\bibitem[{Wang et~al.(2018)Wang, Singh, Michael, Hill, Levy, and
  Bowman}]{wang2018glue}
Alex Wang, Amanpreet Singh, Julian Michael, Felix Hill, Omer Levy, and Samuel~R
  Bowman. 2018.
\newblock Glue: A multi-task benchmark and analysis platform for natural
  language understanding.
\newblock \emph{arXiv preprint arXiv:1804.07461}.

\bibitem[{Wang et~al.(2020)Wang, Wei, Dong, Bao, Yang, and
  Zhou}]{wang2020minilm}
Wenhui Wang, Furu Wei, Li~Dong, Hangbo Bao, Nan Yang, and Ming Zhou. 2020.
\newblock Minilm: Deep self-attention distillation for task-agnostic
  compression of pre-trained transformers.
\newblock \emph{arXiv preprint arXiv:2002.10957}.

\bibitem[{Wenzek et~al.(2020)Wenzek, Lachaux, Conneau, Chaudhary, Guzm{\'a}n,
  Joulin, and Grave}]{wenzek2020ccnet}
Guillaume Wenzek, Marie-Anne Lachaux, Alexis Conneau, Vishrav Chaudhary,
  Francisco Guzm{\'a}n, Armand Joulin, and {\'E}douard Grave. 2020.
\newblock Ccnet: Extracting high quality monolingual datasets from web crawl
  data.
\newblock In \emph{Proceedings of The 12th Language Resources and Evaluation
  Conference}, pages 4003--4012.

\bibitem[{Xu et~al.(2020)Xu, Zhou, Ge, Wei, and Zhou}]{xu2020bert}
Canwen Xu, Wangchunshu Zhou, Tao Ge, Furu Wei, and Ming Zhou. 2020.
\newblock Bert-of-theseus: Compressing bert by progressive module replacing.
\newblock In \emph{Proceedings of the 2020 Conference on Empirical Methods in
  Natural Language Processing (EMNLP)}, pages 7859--7869.

\bibitem[{Zhou et~al.(2020)Zhou, Chelba, Yuezhang, and Li}]{zhou2020practical}
Junpei Zhou, Ciprian Chelba, Yuezhang, and Li. 2020.
\newblock \href {http://arxiv.org/abs/2005.03519} {Practical perspectives on
  quality estimation for machine translation}.

\end{thebibliography}
\bibliographystyle{acl_natbib}

\end{document}